\pdfoutput=1

\documentclass[11pt]{article}

\usepackage[final]{acl}
\usepackage{tabularx}
\usepackage{times}
\usepackage{footnote}
\makesavenoteenv{tabular}
\usepackage{latexsym}
\usepackage{booktabs}
\usepackage{array}
\usepackage{multirow}
\usepackage{enumitem}

\usepackage[T1]{fontenc}
\usepackage{tabularx}

\usepackage[utf8]{inputenc}

\usepackage{microtype}

\usepackage{inconsolata}

\usepackage{graphicx}
\usepackage{hyperref}

\title{The Russian-focused embedders' exploration: ruMTEB benchmark and Russian embedding model design}

\author{
 \textbf{Artem Snegirev},
 \textbf{Maria Tikhonova},
 \textbf{Anna Maksimova},
 \\
 \textbf{Alena Fenogenova},
 \textbf{Alexander Abramov}
\\
 \textsuperscript{}SaluteDevices
\\
 \small{
   \textbf{Correspondence:} \href{mailto:artem.s.snegirev@gmail.com}{artem.s.snegirev@gmail.com}
 }
}

\begin{document}
\maketitle
\begin{abstract}
Embedding models play a crucial role in Natural Language Processing (NLP) by creating text embeddings used in various tasks such as information retrieval and assessing semantic text similarity. This paper focuses on research related to embedding models in the Russian language. It introduces a new Russian-focused embedding model called ru-en-RoSBERTa and the ruMTEB benchmark, the Russian version extending the Massive Text Embedding Benchmark (MTEB). Our benchmark includes seven categories of tasks, such as semantic textual similarity, text classification, reranking, and retrieval.
The research also assesses a representative set of Russian and multilingual models on the proposed benchmark. The findings indicate that the new model achieves results that are on par with state-of-the-art models in Russian. We release the model ru-en-RoSBERTa, and the ruMTEB framework comes with open-source code, integration into the original framework and a public leaderboard.
\end{abstract}

\section{Introduction}
\label{sec:intro}

Text embeddings play an important role in many Natural Language Processing (NLP) tasks, from clustering to semantic textual similarity (STS) and information retrieval (IR). 
The community has addressed this demand by releasing several powerful text embedding models (or embedders)
~\cite{wang2024multilingual, wang2023improving, chen2024bge}. However, there is still a lack of such embedders developed specifically for the Russian language. The most popular Russian-oriented models, such as rubert-tiny2~\footnote{\url{https://huggingface.co/cointegrated/rubert-tiny2}}, SBERT$_{\textnormal{\scriptsize large-nlu-ru}}$\footnote{\url{https://huggingface.co/ai-forever/sbert_large_nlu_ru}}, and SBERT$_{\textnormal{\scriptsize large-mt-nlu-ru}}$\footnote{\url{https://huggingface.co/ai-forever/sbert_large_mt_nlu_ru}}, have been released several years ago and thus do not include modern data in their training corpora. The latest models are based on an outdated version of the ruBERT~\cite{zmitrovich2023family}~\footnote{\url{https://huggingface.co/ai-forever/ruBert-large}} 
model as a backbone. Moreover, being monolingual, they can not profit from knowledge transfer between languages.

\begin{figure}[t]
\vspace{-30pt}
\includegraphics[width=\linewidth,alt={The illustration of the ruMTEB benchmark.}]{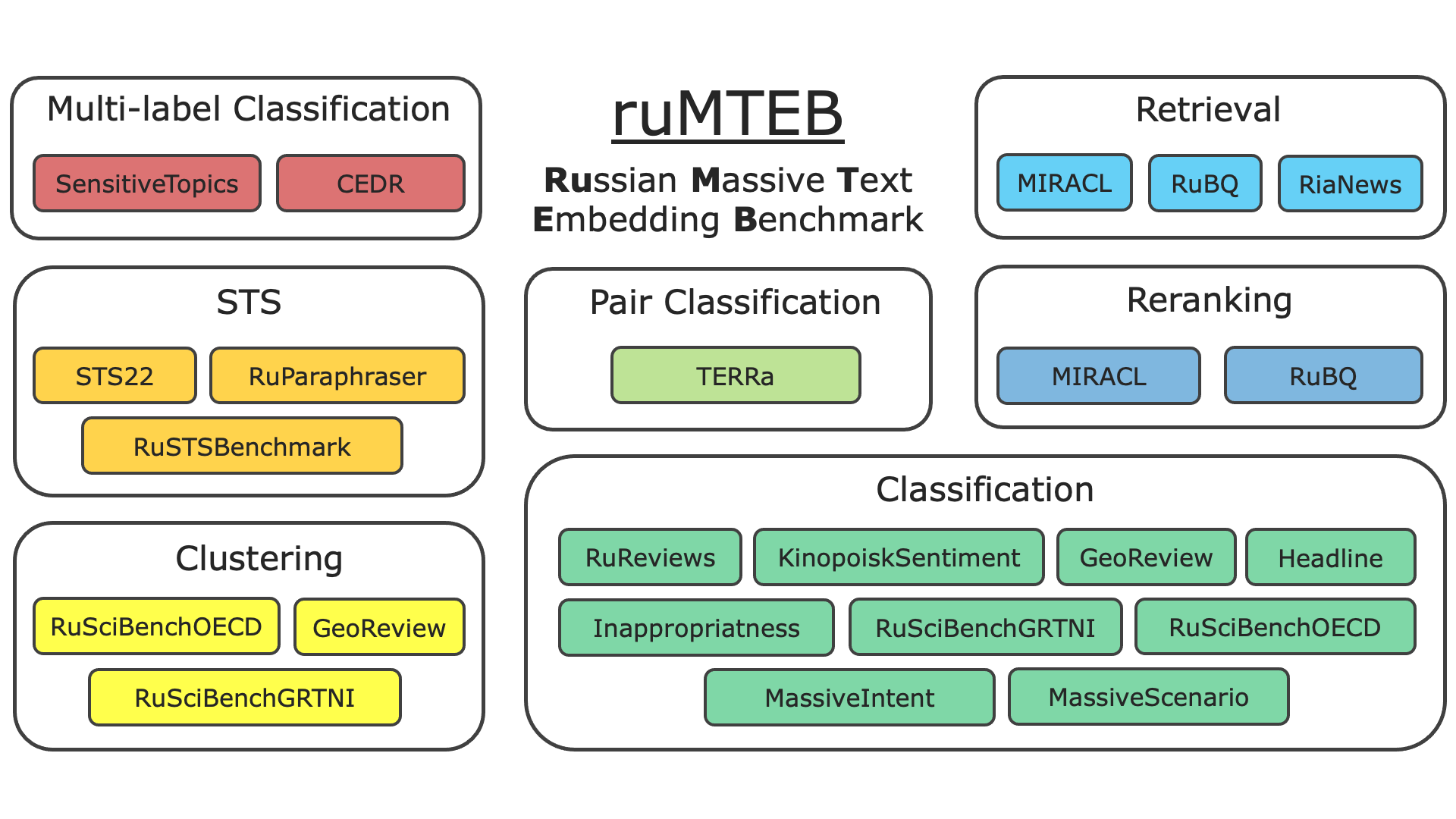}
\caption{The scheme of the ruMTEB benchmark presenting all benchmark tasks divided into 7 task categories.}
\label{fig:mteb_idea}
\end{figure}

Given the usability of such text embeddings, evaluating their quality and the corresponding embedders is also important. One general approach is to evaluate text embeddings on a set of standard text embedding tasks (classification, clustering, etc.) For English, Massive Text Embedding Benchmark, or MTEB~\cite{muennighoff2023mteb}, is considered to be a standard for such an evaluation. For Russian, however, there are significantly fewer evaluation resources. Only few tasks from MTEB contain Russian subsets, while until recently, the only embedding benchmark for Russian was enkodechka~\footnote{\url{https://github.com/avidale/encodechka}}, which appeared several years ago and is still actively used. However, it has significantly fewer tasks than MTEB and has no tasks to evaluate the retrieval abilities of the model.

This paper addresses both problems by presenting a novel Russian-focused embedding model, also adapted for the English language, allowing knowledge transfer from this high resource language, and a new benchmark for text embedding evaluation in Russian called ruMTEB (see Figure~\ref{fig:mteb_idea} for its general structure), which contains 23 text embedding tasks in MTEB format. Among them, 17 tasks are new, and the other 6 are formed on the multilingual MTEB datasets. Thus, the contributions of our work are the following:
\begin{itemize}
    \item we publicly release a Russian-focused text embedding model~\footnote{\url{https://huggingface.co/ai-forever/ru-en-RoSBERTa}}
    adapted for the English language;
    \item we present the Russian version of MTEB and release 17 new Russian datasets for text embedding evaluation~\footnote{\url{https://huggingface.co/collections/ai-forever/ru-mteb-6650a6a6708dc5107a9e0ba3}}, which form the benchmark backbone, and a public leaderboard~\footnote{\url{https://huggingface.co/spaces/mteb/leaderboard}};
    \item we explore model cross-lingual transfer knowledge abilities and various model training hypotheses, which define our final training pipeline;
    \item we evaluate the presented model on ruMTEB and compare its performance with a set of open-source baseline solutions.
\end{itemize}

\section{Related Work}
\label{sec:related_work}
\subsection{Text Embedding Models}
General text embedding models are widely used in various applications such as retrieval-augmented generation (RAG)~\cite{lewis2020retrieval}, STS, as well as multimodal scenarios~\cite{radford2021learning}. One of the first approaches for training such models was to fine-tune a pre-trained language model on the collection of labeled text pairs, such as SNLI~\cite{bowman2015large}. Natural Language Inference (NLI) has been shown~\cite{reimers-2019-sentence-bert} to help such models learn useful representations of texts for STS and other downstream applications. Recent approaches for model training utilize labeled datasets, which can be divided into symmetric (NLI, STS) and asymmetric (Retrieval) tasks. Hence, the training objective takes the form of multitask learning over one or multiple objectives, and the specialized instructions are applied for each task~\cite{INSTRUCTOR}.

Instead of training on limited labeled datasets, in~\cite{wang2022text}, it has been proposed to split fine-tuning into two stages: contrastive pre-training uses a large-scale pair dataset of noisy (or weakly-supervised) text examples, and contrastive fine-tuning utilize a smaller number of high-quality examples. The authors of the E5$_{\textnormal{\scriptsize mistral-7b-instruct}}$~\cite{wang2023improving} utilize an approach for model training which does not include expensive contrastive pre-training that has been shown to be useful for smaller encoder-only model XLM-R~\cite{conneau2019unsupervised}. While their quality remains comparable, encoder-only models are more cost-effective during inference.

Examples of modern English-focused models include E5~\cite{wang2022text}, BGE~\cite{xiao2023c}, GTE~\cite{li2023towards}, Nomic~\cite{nussbaum2024nomic} and Arctic Embed~\cite{merrick2024arctic}. Scaling the number of languages supported (including Russian) has been demonstrated in mE5~\cite{wang2024multilingual} models and BGE-M3~\cite{chen2024bge}, thereby extending their applicability in multilingual contexts. The Russian-oriented models are mainly represented by SBERT models and rubert-tiny2 and their modifications.

Models mentioned above are often used for additional fine-tuning on a specific task. To preserve the general ability of the embedding model, it has been proposed~\cite{xiao2023lm} to merge the fine-tuned model with its base model.

To address this lack of contemporary Russian-focused embedding models performing on par with their multilingual counterparts, we present \textbf{ru-en-RoSBERTa}.

\subsection{Text Embedding Benchmarks}
Model evaluation has always played an inevitable role in NLP progress. Starting from GLUE~\cite{wang2018glue} and SuperGLUE~\cite{wang2019superglue} benchmarks have been a standard model evaluation method. As for text embedding representation evaluation, it has been in focus for many years and was commonly evaluated on a STS, for which yearly released SemEval~\cite{agirre2016semeval, cer2017semeval, chen2022semeval}\footnote{\url{https://semeval.github.io/}} datasets were commonly used. Being a single dataset inevitably limits the SemEval expressivity. Following the same approach, SentEval~\cite{conneau2018senteval}, which focuses on classifier models on top of embedding, overcomes this limitation by aggregating multiple STS datasets. Still, it lacks the evaluation instruments for the suitability of embedding for retrieval or clustering tasks. Due to the inefficiency of a single STS evaluation, USEB~\cite{wang2021tsdae}, focusing on reranking tasks, and BEIR~\cite{thakur2021beir}, aimed at zero-shot information retrieval evaluation, were created. SciRepEval~\cite{singh2023scirepeval}, a multi-format benchmark for scientific document representations, includes 24 realistic tasks across four formats: classification, regression, ranking, and search. Finally, uniting and unifying all main classes of the embedding tasks MTEB~\cite{muennighoff2023mteb} has been proposed and is now considered a multilingual text embedding evaluation standard. Moreover, its approach was also adopted and recreated for the Scandinavian languages in the SEB~\cite{enevoldsen2024scandinavian} and Chinese in the C-MTEB~\cite{xiao2023c}\footnote{\url{https://huggingface.co/C-MTEB}} benchmarks.

Most of the benchmarks mentioned above are English-focused. Even MTEB, despite being multilingual, lacks Russian-language tasks. Only few of the datasets contain Russian subsets, which is not enough for a proper embedding evaluation in this language. Apart from these few MTEB subsets, the only Russian embedding benchmark remains enkodechka, which has significantly fewer tasks than MTEB and no tasks to evaluate the model's retrieval abilities.

Still, there is a need to evaluate text embedding in Russian. To address this demand, we propose \textbf{ruMTEB} comprising a set of text embedding tasks in MTEB format.

\section{ruMTEB Embedding Benchmark}
\label{sec:rumteb}

\subsection{Benchmark Structure and Evaluation Methodology}
\label{subsec:tasks}
The ruMTEB benchmark unites 23 datasets, which can be divided into 7 task categories similar to the corresponding categories in the original MTEB benchmark: Classification (9 datasets), Clustering (3 datasets), MultiLabel Classification (2 tasks), Pair Classification (1 task), Reranking (2 tasks), Retrieval (3 tasks), and STS (3 tasks). 
Below each task category, the evaluation process is briefly described, and the dataset information can be found in Subsection~\ref{subsec:datasets}.

\vspace{0.5em} \noindent\textbf{Classification}.
The evaluation is performed in 10 consecutive experiments (bootstrap evaluation). For run, a bootstrap subset of $n$ (by default, $n=8$) training samples is sampled, and this down-sampled train and test parts are embedded using the embedding model. The training subset is used to train the logistic regression classifier (with 100 interactions maximum). Then, test predictions are scored using the standard Accuracy score.

\vspace{0.5em} \noindent\textbf{Pair Classification}. This group includes datasets where, given a pair of text labels, one has to predict a binary label. For evaluation, the two texts in each pair are embedded via the embedding model, and the cosine similarity between their embeddings is computed. Then, using the best binary threshold, average precision is computed.

\vspace{0.5em} \noindent\textbf{Multi-label Classification}. For evaluation, train and test sets are embedded. Then bootstrap evaluation with 10 runs is performed. In each run the training sets are down-sampled to 8 instances of each unique label. The train embeddings are used to train the kNN classifier (n\_neighbours = 5). The result is evaluated on the test part using the standard Accuracy score.

\vspace{0.5em} \noindent\textbf{Clustering}.
This task type includes datasets where, given a set of text fragments, one has to group them into meaningful clusters.
For evaluation, text fragments are embedded. Then bootstrap evaluation with 10 runs is performed. For each run, a subset of embedding are samples, which are then clustered using K-means clustering. The result is evaluated via v-measure~\cite{rosenberg2007v} and averaged over all experiments.

\vspace{0.5em} \noindent\textbf{Semantic Textual Similarity (STS)}. Given a pair of
sentences, the goal is to determine their textual similarity. Labels are continuous scores ranging from 0 to 1 (the closer to 1, the more similar). For evaluation, cosine similarity over the embedded sentences for each pair is computed. The result is evaluated with Spearman correlation~\cite{reimers2016task}.

\vspace{0.5em} \noindent\textbf{Reranking}.
Inputs are a query and a list of reference texts (relevant and irrelevant). The goal is to correctly rank these texts according to their relevance to the query. For evaluation, the texts for each query are ranked by the cosine similarity between the query embedding and the embedding of the given texts. The obtained ranking is scored with MAP@k ($k = 10$)\footnote{The exception is MIRACLReranking which is evaluated using nDCG@10 following the original MTEB methodology.} for each query and averaged over all queries.

\vspace{0.5em}\noindent\textbf{Retrieval}. For this task type, each dataset includes a set of documents and queries and a mapping for each query to relevant documents. The task aims to find relevant documents for each task. For evaluation, each query document is ranked by the cosine similarity computed between the document embedding and the query embedding. The result is evaluated using nDCG@10.

\subsection{Benchmark Tasks}
\label{subsec:datasets}
\begin{table*}[ht!]
    \setlength{\tabcolsep}{3pt}
    \centering
    \small
    \renewcommand\arraystretch{0.95}
   
    \begin{tabularx}{\textwidth}{%
       l%
        l%
        l%
        >{\raggedright\arraybackslash}X%
        l%
        l%
        l%
        l%
        l@{}%
    }%

        \toprule
         \textbf{Task Category} &  & \textbf{Task name} & \textbf{Data origin}  & \textbf{Train} & \textbf{Val} & \textbf{Test}    \\
        \midrule
        \multirow{9}{*}{\rotatebox[origin=c]{0}{Classification}} &    & GeoReviewClassification & \href{https://github.com/yandex/geo-reviews-dataset-2023}{Geo Reviews} & 50000 & 5000 & 5000  \\
         &  & HeadlineClassification & \href{http://paraphraser.ru/download/get?file_id=7}{ParaPhraserPlus} & 36000 & 12000 & 12000  \\
        &  & InappropriatnessClassification & \href{https://github.com/s-nlp/inappropriate-sensitive-topics/blob/main/Version3/Inappapropriate_messages.csv}{Inappropriate Sensitive Topics} & 4000 & 4000 & 10000 \\
        &          & KinopoiskSentimentClassification & \href{https://huggingface.co/datasets/blinoff/kinopoisk?row=2}{Kinopoisk Movie Reviews}  & 10500 & 1500  & 1500\\

         &          & \textit{MassiveIntentClassification} & \href{https://huggingface.co/datasets/mteb/amazon_massive_intent}{MTEB}  & 11514 & 2033  & 2974\\ 
         &          & \textit{MassiveScenarioClassification} & \href{https://huggingface.co/datasets/mteb/amazon_massive_scenario}{MTEB}  & 11514 & 2033  & 2974\\
         &  & RuReviewsClassification& \href{https://github.com/sismetanin/rureviews}{RuReviews} & 45000 & 15000 & 15000  \\
         &  & RuSciBenchGRTNIClassification & \href{https://huggingface.co/datasets/mlsa-iai-msu-lab/ru_sci_bench}{RuSciBench} & 28476 & -- & 2773 & \\
         &  & RuSciBenchOECDClassification & \href{https://huggingface.co/datasets/mlsa-iai-msu-lab/ru_sci_bench}{RuSciBench} & 27783 & -- & 3220 & \\\midrule
        \multirow{1}{*}{\rotatebox[origin=c]{0}{PairClassification}} 
         &  & TERRa & \href{https://russiansuperglue.com/tasks/task_info/TERRa}{TERRa} & 2616 & 307 & --   \\
         
        \midrule
        \multirow{2}{*}{\rotatebox[origin=c]{0}{MultiLabelClassification}} 
        
         & & CEDRClassification & \href{https://huggingface.co/datasets/sagteam/cedr_v1}{CEDR} & 7529 & -- & 1882   \\
         &  & SensitiveTopicsClassification & \href{https://github.com/s-nlp/inappropriate-sensitive-topics/blob/main/Version3/sensitive_topics.csv}{Inappropriate Sensitive topics} & 29178 & -- & 2048   \\
         
        \midrule
        \multirow{3}{*}{\rotatebox[origin=c]{0}{STS}} &  & RuSTSBenchmarkSTS & \href{https://huggingface.co/datasets/PhilipMay/stsb_multi_mt}{STS Benchmark} & 5224 & 1336 & 1264   \\
         &  & \textit{STS22} & \href{https://huggingface.co/datasets/mteb/sts22-crosslingual-sts}{MTEB} & -- & -- & 265   \\
         &  & \textit{RuParaphraserSTS} & \href{https://huggingface.co/datasets/merionum/ru_paraphraser}{MTEB} & 7227 & -- & 1924   \\
        \midrule
        \multirow{2}{*}{\rotatebox[origin=c]{0}{Clustering}} &  & GeoReviewClustering & \href{https://github.com/yandex/geo-reviews-dataset-2023}{Geo Reviews} & -- & -- & 2000  \\
         &  & RuSciBenchGRTNIClustering & \href{https://huggingface.co/datasets/mlsa-iai-msu-lab/ru_sci_bench}{RuSciBench} & -- & -- & 31080  \\

          &  & RuSciBenchOECDClustering & \href{https://huggingface.co/datasets/mlsa-iai-msu-lab/ru_sci_bench}{RuSciBench} & -- & -- & 30740   \\
        
          \midrule
         \multirow{2}{*}{\rotatebox[origin=c]{0}{Reranking}}
         &  &\textit{MIRACLReranking} & \href{https://huggingface.co/datasets/miracl/mmteb-miracl-reranking}{MTEB}& -- & 44608 & --  \\
          & & RuBQReranking & \href{https://github.com/vladislavneon/RuBQ/tree/master/RuBQ_2.0}{RuBQ 2.0} & -- & -- & 1551   \\

        \midrule
         \multirow{3}{*}{\rotatebox[origin=c]{0}{Retrieval}}          &  & \textit{MIRACLRetrieval} & \href{https://huggingface.co/datasets/miracl/mmteb-miracl}{MTEB}& -- & 13100 & --  \\
        &  & RiaNewsRetrieval & \href{https://github.com/RossiyaSegodnya/ria_news_dataset}{Ria News}& -- & -- & 10000  \\
         &  & RuBQRetrieval & \href{https://github.com/vladislavneon/RuBQ/tree/master/RuBQ_2.0}{RuBQ 2.0} & -- & -- & 2845   \\

        \bottomrule
    \end{tabularx}

\caption{The ruMTEB task outline. The \textbf{Train}, \textbf{Val}, and \textbf{Test} columns show the sizes of the dataset splits (``--'' means the absence of the split). Datasets from the original MTEB benchmark are in Italic; for them, the sizes of the Russian subsets are reported.
    }\label{tab:tasks_outline}
\end{table*}
ruMTEB comprises 23 datasets divided into 7 task types mentioned above: six datasets based on the Russian subsets from the original multilingual MTEB set (MassiveIntendClassification, MassiveScenarioClassification, MIRACLReranking, MIRACLRetrieval, RuParaphraserSTS, STS22) and 17 new datasets we release within the research. The latter are based on popular Russian time-tested and community-tested datasets.

We took the datasets based on the original MTEB without any changes. For the Russian community datasets, we selected only the tests with high-quality labeling, relying on the original publications. We performed data cleaning and automatic filtering where necessary, removed duplicates, manually verified small subsets of examples, and formatted them in the MTEB format. The main dataset information and their statics are given in Table~\ref{tab:tasks_outline}, and the detailed task descriptions and preprocessing for the new sets are in Appendix~\ref{app:data_prep}.

\section{Text Embedding Model for Russian}
\label{sec:text_embedding_model}

This section is devoted to the text embedding model \textbf{ru-en-RoSBERTa} released within the research. We describe the training data, the base model, and the final training pipeline, motivated by the experiments described in Section~\ref{sec:analysis}.

\subsection{Training Data}
\label{subsec:text_embedding_model__training_data}

Following previous work~\cite{wang2022text, li2023towards, nussbaum2024nomic}, we use publicly available data, high-quality and synthetic datasets to create training pairs (see Appendix~\ref{app:dataset_info} for the full training list)~\footnote{To avoid potential data leakage we use only the training parts of all the sets.}, which, for experiment purpose (see Section~\ref{sec:analysis}), we divide into four groups described below.

\vspace{0.5em} \noindent\textbf{Basic Russian Datasets}. This group consists of 17 tasks. It includes pairs from SberQuAD~\cite{sberquad}, XNLI~\cite{conneau2018xnli}, parallel translations~\cite{banon-etal-2020-paracrawl,tiedemann-2012-parallel,zhang-etal-2020-improving}, and publicly available data from various domains, such as news, blogs, QA platforms, and other Internet resources. We filter this data mostly with manual rules (see Appendix~\ref{app:data_filtration} for the details).

\vspace{0.5em} \noindent\textbf{Basic English Datasets}. The group is formed from MEDI~\cite{INSTRUCTOR} corpus without provided instructions. We also exclude instructional datasets from Super-NI~\cite{wang2022super} and thus retain 30 datasets representing different domains and tasks. We do not apply any additional preprocessing steps.

\vspace{0.5em} \noindent\textbf{Additional Synthetic Datasets}. The group includes Query2doc MS-MARCO~\cite{wang2023query2doc}, DINO-STS-x1x2~\cite{schick2020generating}, RuHNP~\cite{deepvk2024ru_hnp}, entailment and contradiction pairs from RuWANLI~\cite{deepvk2024ru_wanli}, and a sample of generated pairs by ruT5-base\footnote{\url{https://huggingface.co/ai-forever/ruT5-base}} model from WikiOmnia~\cite{pisarevskaya-shavrina-2022-wikiomnia}. We do not change the data content and use the datasets as is.

\vspace{0.5em} \noindent\textbf{Additional Retrieval Datasets}. We use Russian and English parts of Mr. Tydi~\cite{mrtydi} and MIRACL~\cite{zhang2022making} from BGE-M3 fine-tuning data. These datasets provide high-quality examples and are designed for the same retrieval tasks included in our benchmark.

We mine negatives similar to~\cite{xiao2023c} using the mE5$_{\textnormal{\scriptsize small}}$~\cite{wang2024multilingual} and sample documents by rank in the range of 20-100. For all datasets, the provided hard negatives are also used. For additional synthetic and retrieval datasets, the provided negatives are used (if available), and the rest are randomly sampled from the same dataset.

\begin{table*}[h]
\small
    \centering
    \begin{tabular}{l*{8}{c}}
        \toprule
        \textbf{Data Source}   & \textbf{Cls.}   & \textbf{Clust.}   & \textbf{MultiLabelCls.}   & \textbf{PairCls.}   & \textbf{Rerank.}   & \textbf{Retr.}   & \textbf{STS}   & \textbf{Avg.}   \\
        \midrule
        Basic English Datasets       & 61.7   & \textbf{56.6}     & 36.8             & 54.7       & 57.5      & 57.6    & 69.9  & 56.4  \\
        Basic Russian Datasets      & 60.0   & 54.2     & \underline{38.0}             & 56.3       & 60.8      & 61.7    & 72.3  & 57.6  \\
        Mixture       & 61.4   & 54.3     & 37.8             & 56.5       & 61.4      & 63.8    & 72.9  & 58.3  \\
        \quad + synthetic   & \textbf{62.3}   & \underline{54.6}     & \textbf{39.0}             & \underline{59.7}       & \underline{62.1}      & \underline{64.1}    & \underline{73.6}  & \underline{59.3}  \\
        \quad + synthetic \& retrieval   & \underline{62.1}   & 53.9     & \textbf{39.0}             & \textbf{60.0}       & \textbf{63.1}      & \textbf{65.1}    & \textbf{73.7}  & \textbf{59.6}  \\
        \bottomrule
    \end{tabular}
    \caption{Different data sources impact. Model performance is measured on ruMTEB. \textbf{Avg.} stands for the average score and is computed as the mean of the category scores. The best score is put in bold, the second best is underlined.}
    \label{tab:analysis_data_sources}
\end{table*} 

\subsection{Base Model and English Language Adaptation}
\label{subsec:text_embedding_model__base_model}

Since we focus on the Russian language, we use ruRoBERTa~\footnote{\url{https://huggingface.co/ai-forever/ruRoberta-large}}~\cite{zmitrovich2023family}, which has the highest scores on the classic Russian SuperGLUE~\cite{shavrina2020russiansuperglue} benchmark among the models of its size. In addition, we adapt it to the English language, allowing knowledge transfer from this high-resource language (see Section~\ref{sec:analysis} for the corresponding experiments). 

We extend the original ruRoBERTa tokenizer with tokens from RoBERTa~\footnote{\url{https://huggingface.co/FacebookAI/roberta-large}}~\cite{liu2019roberta}. To learn new token embeddings, we train the model using Masked Language Modeling (MLM) objective~\cite{devlin2018bert}. We use the same hyperparameters as in the ruRoBERTa and the batch size of 1024. We use unique training texts from Section~\ref{subsec:text_embedding_model__training_data} and train for one epoch (\textasciitilde11k steps). 

The whole process takes one day on two A100 80GB cards. To reduce the effect of catastrophic forgetting~\cite{kirkpatrick2017overcoming}, we merge encoder layers using spherical linear interpolation (SLERP) algorithm\footnote{\url{https://gist.github.com/dvschultz/3af50c40df002da3b751efab1daddf2c}} with the factor of 0.25 to the original model.
In our work, we refer to the obtained model version with the extended vocabulary as ru-en-RoBERTa.

\subsection{Contrastive Fine-tuning}
\label{subsec:text_embedding_model__fine_tuning}

Following~\cite{INSTRUCTOR}, we perform contrastive fine-tuning for ru-en-RoBERTa on a mix of supervised and unsupervised data (from the Section~\ref{subsec:text_embedding_model__training_data}). 
We use prefix strategy from ~\cite{cohere_embed_v3} applying prefixes for each pair to avoid a conflicting reward signal (see Appendix~\ref{app:dataset_info} for the prefix rules and the full prefix list).

We employ the standard InfoNCE contrastive loss~\cite{oord2018representation}, keep a fixed temperature value of 0.02, and obtain normalized text embedding using CLS pooling. The batch is filled with pairs of the same dataset (stratified sampling), and proportional batch sampling is applied. Negative examples are formed from 7 hard negatives per query, and the remaining negatives are taken from a batch of the same device (in-batch negatives). After fine-tuning, the SLERP merging is applied to the base model with a factor of 0.1. See \ref{app:training_details} for the training details. We report the computational, energy, and carbon costs in Section~\ref{sec:energy}.

\section{Training Procedure Analysis}
\label{sec:analysis}

This section describes experiments we conducted to determine the final training pipeline. We used basic Russian, English, and additional synthetic datasets, the training approach described in Section~\ref{sec:text_embedding_model} unless otherwise specified, and the full ruMTEB version for evaluation. Details on the model configurations in these experiments are given in~\ref{app:ablation_training} and further findings are in Appendix~\ref{app:additional_notes}.

\subsection{Cross-lingual Knowledge Transfer and Data Sources}
\label{subsec:en_adapta}

We explored five training data configurations to study whether the model can profit from knowledge transfer between languages and various data sources.
For this, we trained embedding models based on ru-en-RoBERTa: on basic English datasets only, basic Russian datasets only, and their mixture, simple or augmented with additional synthetic/synthetic+retrieval datasets. Each model is trained for 1500 steps.

Results presented in Table~\ref{tab:analysis_data_sources} indicate that the embedding model gets better results when trained on data in Russian and English simultaneously. 
Additional synthetic datasets and high-quality retrieval datasets further improve the embedding model quality despite the tasks these datasets solve already being well represented in the basic datasets. Given that in all scenarios, the number of steps is fixed, the change in the results could not account for longer training.

The model especially benefits from synthetic datasets on STS-related tasks, while quality degradation in clustering tasks remains unclear. Note that the model trained on almost all data (except the additional retrieval dataset is better by only 0.6 points. The results obtained on data in English may be due to the better quality of the tasks presented in MEDI.

\subsection{English Language Adaptation}
Having shown that the model can profit from cross-lingual knowledge transfer, we turned to selecting the optimal language adaptation strategy. Namely, we compared:
\begin{itemize}
    \item \textit{ruRoBERTa} and \textit{XLM-R} used as baselines;
    \item \textit{ru-en-RoBERTa} from subsection Section~\ref{subsec:text_embedding_model__base_model};
    \item \textit{ru-en-RoBERTa w/ RetroMAE} same approach as previous where we substituted MLM with RetroMAE~\cite{RetroMAE}, which proved beneficial for BGE-M3 in~\cite{chen2024bge}. For this configuration, set the masking ratio of decoder input tokens to 30\%;
    \item \textit{ru-en-RoBERTa w/o SLERP} same as \textit{ru-en-RoBERTa} without SLERP after English language adaptation.
    
\end{itemize}
We perform contrastive fine-tuning for each model and then evaluate them on ruMTEB. Results (see Table~\ref{tab:analysis_summary}) show that ru-en-RoBERTa outperforms both baselines by a significant margin. Additionally, the fact that XLM-R slightly outperforms ruRoBERTa may indicate that XLM-R copes better with knowledge transfer from basic English datasets, which provide diverse examples of high quality. Merging the encoder layers after language adaptation with the original model improves the model quality while using RetroMAE leads to decreased results. 

\begin{table*}[h]
    \centering
    \small
    \begin{tabular}{l*{8}{c}}
    \toprule
                                  & \textbf{Cls.}    & \textbf{Clust.}  & \textbf{MultiLabelCls.} & \textbf{PairCls.} & \textbf{Rerank.} & \textbf{Retr.}   & \textbf{STS}     & \textbf{Avg.}   \\
    \midrule
    \textit{English Language Adaptation} \\
    \midrule
    XLM-R                         & \textbf{63.0}    & \textbf{56.6}    & 38.7                    & 59.6              & 60.7             & 62.6             & \underline{73.9} & 59.3  \\
    ruRoBERTa                     & 61.4             & 55.8             & 38.5                    & 59.1              & 61.1             & 63.1             & 73.6             & 58.9  \\
    ru-en-RoBERTa\textsuperscript{\textdagger}                 
                                  & 62.5             & 55.8             & \underline{39.1}        & 60.0              & 62.8             & \underline{65.3} & 73.6             & \textbf{59.9}  \\
    \quad w/ RetroMAE             & 62.2             & 55.7             & 37.9                    & 59.1              & 60.1             & 61.9             & 72.7             & 58.5  \\
    \quad  w/o SLERP              & 62.2             & 55.3             & 38.8                    & 59.9              & \underline{62.9} & 64.9             & 73.3             & \underline{59.6}  \\
    \midrule
    \textit{Training Objective} \\
    \midrule
    Additive margin               & 62.4             & 55.3             & 38.9                    & \textbf{60.9}     & 62.7             & 65.2             & 73.7             & \textbf{59.9}  \\
    Document penalty              & 62.5             & \underline{55.9} & \textbf{40.0}           & \textbf{60.9}     & 61.2             & 62.4             & \textbf{74.1}    & \underline{59.6}  \\
    AnglE similarity              & 62.3             & \underline{55.9} & \underline{39.1}        & \underline{59.9}  & 61.8             & 63.7             & 72.7             & 59.3  \\
    Mean pooling                  & 62.6             & 55.5             & 38.4                    & 59.2              & 61.1             & 63.1             & 72.5             & 58.9  \\
    \midrule
    \textit{Training Examples} \\
    \midrule
    Increase hard negatives group & 62.7             & 55.6             & 38.4                    & \underline{60.6}  & \textbf{63.1}    & \textbf{65.7}    & 73.5             & \textbf{59.9}  \\
    Disable stratified sampling   & \underline{62.8} & 55.7             & 38.4                    & 60.3              & 61.2             & 63.0             & 72.6             & 59.2  \\
    Remove prefixes               & 61.5             & 54.4             & 38.1                    & \textbf{60.9}     & 61.4             & 64.1             & 73.4             & 59.1  \\
    \bottomrule
    \end{tabular}
    
    \caption{Results of the model, method, and data variation. \textbf{Avg.} is the average of the category results.{}\textsuperscript{\textdagger}The reference results for the training objective and training examples sections is model based on ru-en-RoBERTa. Each experiment changes a single component (e.g., use AnglE similarity instead of cosine). Model performance is evaluated on ruMTEB. The best score across all experiments is bold, the second best is underlined.}\label{tab:analysis_summary}
\end{table*}

\subsection{Training Examples}
In this series of experiments (see Table~\ref{tab:analysis_summary}), we show the effects of prefixes, stratified sampling, and the number of hard negatives. 

\vspace{0.5em} \noindent\textbf{Remove prefixes}. Fine-tuning the model on symmetric and asymmetric tasks simultaneously can hurt performance without instructions but improve it when instructions are used~\cite{INSTRUCTOR}. We found that removing prefixes consistently worsens the results, but STS-related tasks were not as affected.

\vspace{0.5em} \noindent\textbf{Disable stratified sampling}. To explore whether stratified sampling is beneficial~\cite{merrick2024arctic} in our case, we disabled it, used prefixes only for queries, and negatives were exchanged across devices. The latter increases the number of negatives per query to 8k. We found that the stratified version works better.

\vspace{0.5em} \noindent\textbf{Hard negatives}. To study whether adding more hard negatives~\cite{ren2021rocketqav2} is beneficial, we increased their number to 15 and reduced per device batch size to 64, maintaining the total number of negative examples. To keep the same number of steps, we apply gradient accumulation. Similarly to~\cite{nussbaum2024nomic}, we found that despite processing almost twice as many texts, the results did not improve. 

\subsection{Training Objective} 
In this experiment (see Table~\ref{tab:analysis_summary}), we examine four modifications of the training objective described in Section~\ref{subsec:text_embedding_model__fine_tuning}.

\vspace{0.5em} \noindent\textbf{Additive margin}. Following~\cite{yang2019improving}, we applied an additive margin with the value of 0.01, and larger values caused convergence problems. We do not use additive margin in our final model and found that datasets are sensitive to margin values. 

\vspace{0.5em} \noindent\textbf{Document penalty}. The authors of GTE~\cite{li2023towards} add a penalty for query-query, document-document, and document-query matching in the denominator of InfoNCE loss that improves model performance on MTEB. We applied a similar approach to~\cite{yang2019improving, INSTRUCTOR}, adding a penalty for document-query matching as an additional loss. The penalty significantly improved model performance on STS and worsened on Retrieval.

\vspace{0.5em} \noindent\textbf{AnglE similarity}. Normalized dot-product is usually used to score a pair of texts. AnglE similarity was proposed in~\cite{li2023angle} to optimize the angle difference of pairs in complex space. We replaced cosine similarity with AnglE. It is worth noting that in the original work AnglE was used in slightly different scenario, therefore, not finding any improvement, we left this for future work.

\vspace{0.5em} \noindent\textbf{Mean pooling}. Without an experiment, the choice of pooling strategy remains unclear. Mean pooling is used in E5, GTE and Nomic~\cite{li2023towards, nussbaum2024nomic}, BGE and Arctic Embed~\cite{xiao2023c, merrick2024arctic} are apply CLS pooling. We observed consistent improvement of the latter, compared to mean pooling.
\begin{table*}[]
\centering
\small
\renewcommand{\arraystretch}{1.05}
\begin{tabular}{llll}
\hline

\textbf{Model Name}                                                  & \textbf{\textbf{Para\-meters}} & \textbf{HuggingFace Hub Link} & \textbf{Citation} \\ \hline

rubert-tiny2   & 29.4M                    & \href{https://huggingface.co/cointegrated/rubert-tiny2}{cointegrated/rubert-tiny2}              & -  \\

SBERT$_{\textnormal{\scriptsize large-nlu-ru}}$      & 427M                     & \href{https://huggingface.co/ai-forever/sbert_large_nlu_ru}{ai-forever/sbert\_large\_nlu\_ru}            &  - \\ 

SBERT$_{\textnormal{\scriptsize large-mt-nlu-ru}}$         & 427M                     & \href{https://huggingface.co/ai-forever/sbert_large_mt_nlu_ru}{ai-forever/sbert\_large\_mt\_nlu\_ru}            & - \\ 

ru-en-RoSBERTa   & 404M                    & \href{https://huggingface.co/ai-forever/ru-en-RoSBERTa}{ai-forever/ru-en-RoSBERTa}              & -  \\

mE5$_{\textnormal{\scriptsize small}}$                              & 118M                     & \href{https://huggingface.co/intfloat/multilingual-e5-small}{intfloat/multilingual-e5-small}             & \citet{wang2024multilingual}   \\

mE5$_{\textnormal{\scriptsize base}}$                              & 278M                     & \href{https://huggingface.co/intfloat/multilingual-e5-base}{intfloat/multilingual-e5-base}             & \citet{wang2024multilingual}  \\

mE5$_{\textnormal{\scriptsize large}}$                              & 560M                     & \href{https://huggingface.co/intfloat/multilingual-e5-large}{intfloat/multilingual-e5-large}             & \citet{wang2024multilingual}  \\

BGE-M3 & 567M & \href{https://huggingface.co/BAAI/bge-m3}{BAAI/bge-m3}& \citet{multim3}\\

mE5$_{\textnormal{\scriptsize large-instruct}}$                              & 560M                     & \href{https://huggingface.co/intfloat/multilingual-e5-large-instruct}{intfloat/multilingual-e5-large-instruct}             & \citet{wang2024multilingual}  \\

E5$_{\textnormal{\scriptsize mistral-7b-instruct}}$ & 7.11B & \href{https://huggingface.co/intfloat/e5-mistral-7b-instruct}{intfloat/e5-mistral-7b-instruct}& \citet{wang2023improving}\\

\bottomrule
\end{tabular}
\caption{The evaluated mode description. Instruct models are marked with the corresponding suffix.}
\label{tab:model_baselines}
\end{table*}


\begin{table*}[h]
    \centering
    \small
    \renewcommand{\arraystretch}{1.05}
    \begin{tabular}{l*{8}{c}}
    \toprule
    \textbf{Model name}                              & \textbf{Cls.}   & \textbf{Clust.}   & \textbf{MultiLabelCls.}   & \textbf{PairCls.}   & \textbf{Rerank.}   & \textbf{Retr.}   & \textbf{STS}   & \textbf{Avg.}   \\
    \midrule
    rubert-tiny2                                         & 52.17             & 39.12             & 29.45             & 51.87             & 30.95             & 8.89              & 61.60             & 42.22             \\
    SBERT$_{\textnormal{\scriptsize large-nlu-ru}}$         & 57.24             & 50.44             & 31.87             & 50.17             & 32.81             & 8.51              & 57.21             & 45.35             \\
    SBERT$_{\textnormal{\scriptsize large-mt-nlu-ru}}$      & 57.52             & 51.29             & 32.67             & 51.97             & 40.56             & 19.13             & 64.40             & 48.72             \\
    mE5$_{\textnormal{\scriptsize small}}$               & 56.44             & 51.35             & 31.99             & 55.14             & 65.28             & 65.85             & 69.48             & 57.29             \\
    mE5$_{\textnormal{\scriptsize base}}$                & 58.26             & 50.27             & 33.65             & 54.98             & 66.24             & 67.14             & 70.16             & 58.34             \\
    mE5$_{\textnormal{\scriptsize large}}$               & 61.01             & 52.23             & 36.00             & 58.42             & 69.65             & 74.04             & 71.62             & 61.41             \\
    BGE-M3                                               & 60.46             & 52.38             & 34.86             & 60.60             & \underline{69.71} & \textbf{74.79}    & 73.68             & 61.58             \\
    ru-en-RoSBERTa                                       & 62.74             & 56.06             & 38.88             & 60.79             & 63.89             & 66.52             & \underline{73.97} & 61.77             \\
    \midrule
    mE5$_{\textnormal{\scriptsize large-instruct}}$      & \underline{66.31} & \underline{63.21} & \underline{41.15} & \textbf{63.89}    & 69.17             & \underline{74.41} & \textbf{74.85}    & \underline{66.03} \\
    E5$_{\textnormal{\scriptsize mistral-7b-instruct}}$  & \textbf{69.11}    & \textbf{64.24}    & \textbf{42.93}    & \underline{60.81} & \textbf{69.96}    & 74.19             & 73.71             & \textbf{67.18}    \\
    \bottomrule
    \end{tabular}
    \caption{Average model results on ruMTEB task categories. The result for each category represents the mean model score on the tasks from the corresponding task types. \textbf{Avg.} stands for the average score and is computed as the mean of the task scores. The best score is put in bold, the second best is underlined.}
    \label{tab:rumteb_category_results}
\end{table*}

\section{Evaluation}
We evaluate \textit{ru-en-RoSBERTa} and 9 publicly available embedding models for Russian, including the multilingual ones and the two instruct models, on the ruMTEB benchmark. See Table~\ref{tab:model_baselines} for the baseline information and Appendix~\ref{app:prompts} for other details. 

\label{subsec:experiments}
We evaluate all models in the same environments
and scenarios by the procedure described in~\ref{subsec:tasks}. We use MTEB framework\footnote{\url{https://github.com/embeddings-benchmark/mteb/tree/1.14.12}} for evaluation where we integrated evaluation on the new ruMTEB tasks~\footnote{\url{https://github.com/embeddings-benchmark/mteb/pull/815}}\footnote{\url{https://github.com/embeddings-benchmark/mteb/pull/881}}.

\section{Results}
\label{sec:results}
Table~\ref{tab:rumteb_category_results} shows model scores averaged within the task category, and detailed results of the task-wise model evaluation are in Appendix~\ref{app:all_results}\footnote{This results are valid for 09.10.2024. Please, refer to the leaderboard at \url{https://huggingface.co/spaces/mteb/leaderboard} for the latest results.}.

Results analysis reveals that there is a gap between instruct and non-instruct models, mE5$_{\textnormal{\scriptsize large-instruct}}$ and E5$_{\textnormal{\scriptsize mistral-7b-instruct}}$ are better than their non-instruct competitors in all task categories except for Retrieval where BGE-M3 is heading the list.
Moreover, while the instruct/non-instruct difference is not that significant for STS and retrieval, the advantage of the instruct models becomes obvious for other task categories.

As for the non-instruct model analysis, it can be seen that BGE-M3, ru-en-RoSBERTa, and mE5$_{\textnormal{\scriptsize large}}$ perform practically on par.
Moreover, ru-en-RoSBERTa performs better than its non-instruct competitors on all task categories except for 2 (Retrieval and Reranking), probably due to the absence of the contrastive pre-train. For Retrieval and Reranking, BGE-M3 and mE5$_{\textnormal{\scriptsize large}}$ receive much better scores, resulting in ru-en-RoSBERTa being in the second place. The evaluation results show that ru-en-RoSBERTa is a robust embedding model suitable for various textual tasks. Additionally, unlike monolingual Russian models, the bilingual nature of ru-en-RoSBERTa allows it to be further trained or fine-tuned using the much more considerable amount of English data available.

The evaluation results positively characterize the benchmark as being complex enough for modern embedding models, allowing researchers to evaluate text embedding at a high level.

\section{Conclusion}
This paper introduces a new Russian-focused embedding model, which also supports English, and a new benchmark for text embedding evaluation, comprising 23 datasets divided into 7 task types. Among the benchmark datasets, 17 datasets are new and were created within this research.

We report the new embedding model architecture design, pre-training corpus, and training procedure details. We describe the datasets comprising the benchmark and propose the methodology for the text embedding evaluation on it inspired by the MTEB benchmark. We evaluate the presented encoding model and several baselines, thus verifying the ruMTEB complexity and performing the comparative analysis of our model results with the results of standard encoders.

\section{Limitation}
\label{sec:limitations}

\vspace{0.5em} \noindent\textbf{Model limitations.}
The training data for ru-en-RoSBERTa includes large segments from the Internet domain. Consequently, it contains various stereotypes and biases from English and Russian sources. Therefore, a proper model evaluation is still needed to explore their possible vulnerabilities in generalizing to out-of-domain data. The model's context is limited to a length of 512.
One of the model's limitations is that due to limited computational resources, we skip the contrastive pre-training stage, leaving it for future work, although it was found~\cite{wang2022text,wang2023improving} to improve the results on the retrieval-related tasks.

\vspace{0.5em} \noindent\textbf{Lack of evaluation in English.} 
In this work, we focus on the Russian language, and therefore, we do not conduct ru-en-RoSBERTa evaluation in English as this is beyond the scope of this work and quite resource-consuming. Nevertheless, we acknowledge that evaluating the model on the machine translation task or on the English data (e.g., the full MTEB benchmark) is valuable. We leave this to future work.

\vspace{0.5em} \noindent\textbf{Speed and optimization.}
The ruMTEB benchmark comprises 23 tasks, including 6 tasks from the multilingual version. As the project is collaborative and we plan to expand the benchmark with new representative tasks, this may lead to resource-intensive and time-consuming runs. Additionally, continuously updating the benchmark makes previous model results obsolete. Due to the potential expansion of ruMTEB with new tasks and the general trend toward using larger models, there is a need to optimize the benchmark evaluation procedure.

\vspace{0.5em} \noindent\textbf{Datasets.}
The collaborative nature and aggregation of the existing sets in the benchmark make it challenging to ensure uniformly high data quality across all tasks. 
For all benchmark datasets we checked the licenses and filtered the datasets. Unfortunately, despite the joint effort, some tasks still possess errors (e.g., incorrect labels for some examples, grammatical errors, surplus technical symbols, etc.). Moreover, there may still be biases in the data across different domains and sources, and there is still a need to extend tasks in some categories. We encourage researchers to collaborate further to fill the gaps and ensure a more comprehensive and balanced language and task representation in the benchmark.

\vspace{0.5em} \noindent\textbf{Data leakage.}
All benchmark datasets are either publicly available or created using data found on the Web. This can lead to data leakage when some models trained on parts of the dataset may produce inflated scores on the benchmark. In the future, it's crucial to develop methods for automatically identifying data leakage in the task.

\section{Ethical Considerations}
\label{sec:ethical}

\vspace{0.5em} \noindent\textbf{Inference Costs.} Evaluating embedding models on ruMTEB depends on its architecture and size and can be optimized with distributed inference libraries. For example, one run of ru-en-RoSBERTa of the complete evaluation experiment on a single A100 GPU 80GB takes approximately 19 hours.

\vspace{0.5em} \noindent\textbf{Energy Efficiency and Usage.}\label{sec:energy}
We compute the $CO_2$ emissions from pre-training and fine-tuning ru-en-RoSBERTa as Equation~\ref{eq:co2}~\cite{strubelletal2019energy}:

\vspace{-2pt}
\begin{equation}
\label{eq:co2}
    CO_2 = \frac{PUE * kWh * I^{CO2}}{1000}
\end{equation}

\noindent The power usage effectiveness ($PUE$) of our data centers is $1.3$. The resulting CO2 emission is 3.66k kg. Model compression techniques can reduce the computational costs associated with model inference.

\vspace{0.5em} \noindent\textbf{Potential Misuse.} The ruMTEB can be used as training data for acceptability classifiers, potentially improving the quality of generated texts. We acknowledge that these improvements in text generation might lead to the misuse of LLMs for harmful purposes. The intended use of ruMTEB is for \textit{research and development purposes}, and we are aware of the potential negative uses.

\vspace{0.5em} \noindent\textbf{AI-assistants Help.} We improve and proofread the text of this paper using Grammarly\footnote{\url{https://app.grammarly.com/}} to correct grammatical, spelling, and style errors and paraphrasing sentences. Thus, some segments of our publication can be potentially detected as AI-generated, AI-edited, or human-AI-generated.

\section*{Acknowledgments} The authors express their sincere gratitude to all those who contributed to the success of the ruMTEB project. In particular, the authors would like to thank David Dale for his collaboration and valuable insights. The authors also acknowledge Ilya Gusev for his significant contributions to the dataset creation. Special thanks are extended to Roman Solomatin for his essential role in organizing the leaderboard.

\bibliography{custom}

\appendix

\section{Appendix}
\label{sec:appendix}
\subsection{Training Data Details}
\subsubsection{Training Data Information}\label{app:dataset_info}
The list of datasets included in ru-en-RoSBERTa training data and the corresponding prefix used for them are given in Table~\ref{tab:dataset_info}.

We use the following basic rules to choose a prefix:

\begin{itemize}
    \item \verb|search_query| and \verb|search_document| prefixes are for answer or relevant paragraph retrieval
    \item \verb|clustering| prefix is for asymmetric retrieval of title or summary and relevant document
    \item \verb|classification| prefix is for symmetric paraphrasing related tasks (STS, NLI, bitext mining)
\end{itemize}

\begin{table*}[h]
    \begin{minipage}{\textwidth}
    \centering
    \small
    \begin{tabular}{*{4}{l}}
        \toprule
        Dataset & Target task & \# of pairs (K) & Prefix type \\
        \midrule
        \textit{Basic English Datasets}~\cite{INSTRUCTOR}\\
        \midrule
        AGNews & Clustering & 45.0 & clustering \\
        AmazonQA & Retrieval & 100.0 & search\_query/search\_document \\
        AmazonReviews & Clustering & 100.0 & clustering \\
        CCNews & Clustering & 25.0 & clustering \\
        CodeSearchNet & Clustering & 15.0 & clustering \\
        ELI5 & Retrieval & 25.0 & search\_query/search\_document \\
        Fever & Retrieval & 75.0 & search\_query/search\_document \\
        Flickr30k & STS & 25.0 & classification \\
        Gooaq & Retrieval & 25.0 & search\_query/search\_document \\
        HotpotQA & Retrieval & 40.0 & search\_query/search\_document \\
        MedMCQA & Retrieval & 30.0 & search\_query/search\_document \\
        MSMARCO & Retrieval & 175.0 & search\_query/search\_document \\
        AllNLI & NLI & 50.0 & classification \\
        NPR & Clustering & 25.0 & clustering \\
        NQ & Retrieval & 50.0 & search\_query/search\_document \\
        PAQ & Retrieval & 25.0 & search\_query/search\_document \\
        PubMed & Clustering & 30.0 & clustering \\
        S2ORC Title-Abstract & Clustering & 100.0 & clustering \\
        SimpleWiki & STS & 5.0 & classification \\
        SPECTER & STS & 50.0 & classification \\
        SQuAD & Retrieval & 25.0 & search\_query/search\_document \\
        StackExchange Duplicates & STS & 25.0 & classification \\
        Trex & Retrieval & 30.0 & search\_query/search\_document \\
        TriviaQA & Retrieval & 50.0 & search\_query/search\_document \\
        WikiAnswers & STS & 25.0 & classification \\
        WikiHow & Clustering & 25.0 & clustering \\
        WoW & Retrieval & 5.0 & search\_query/search\_document \\
        XSUM & Clustering & 30.0 & clustering \\
        Yahoo Title-Answer & Retrieval & 10.0 & search\_query/search\_document \\
        ZeroshotRE & Retrieval & 15.0 & search\_query/search\_document \\
        \midrule
        \textit{Basic Russian Datasets}~\footnote{For non-cited datasets please refer to~\ref{app:data_filtration}} \\
        \midrule
        HabrQnA QA & Retrieval & 100.0 & search\_query/search\_document \\
        HabrQnA Title-Body & Clustering & 100.0 & clustering \\
        Habr Title-Abstract & Clustering & 50.0 & clustering \\
        Paraphrases & STS & 15.0 & classification \\
        MIRACL Title-Paragraph~\cite{zhang2022making} & Clustering & 100.0 & clustering \\
        MultiParaCrawl~\cite{banon-etal-2020-paracrawl} & Bitext Mining & 300.0 & classification \\
        NewsCommentary~\cite{tiedemann-2012-parallel} & Bitext Mining & 25.0 & classification \\
        paraphrase-NMT-Leipzig & STS & 210.0 & classification \\
        OPUS-100~\cite{zhang-etal-2020-improving, tiedemann-2012-parallel} & Bitext Mining & 175.0 & classification \\
        Pikabu Title-Body & Clustering & 100.0 & clustering \\
        RuNews Title-Body & Clustering & 100.0 & clustering \\
        SberQuAD~\cite{sberquad} & Retrieval & 45.0 & search\_query/search\_document \\
        StackOverflow QA & Retrieval & 100.0 & search\_query/search\_document \\
        StackOverflow Title-Body & Clustering & 75.0 & clustering \\
        XNLI~\cite{conneau2018xnli} & NLI & 125.0 & classification \\
        YandexQ QA & Retrieval & 100.0 & search\_query/search\_document \\
        YandexQ Title-Body & Clustering & 55.0 & clustering \\
        \midrule
        \textit{Additional Synthetic Datasets} \\
        \midrule
        DINO-STS-x1x2~\cite{schick2020generating} & STS & 13.0 & classification \\
        Query2doc~\cite{wang2023query2doc} & Retrieval & 500.0 & search\_query/search\_document \\
        RuHNP~\cite{deepvk2024ru_hnp} & STS & 100.0 & classification \\
        RuWANLI~\cite{deepvk2024ru_wanli} & NLI & 34.0 & classification \\
        WikiOmnia~\cite{pisarevskaya-shavrina-2022-wikiomnia} & Retrieval & 100.0 & search\_query/search\_document \\
        \midrule
        \textit{Additional Retrieval Datasets} \\
        \midrule
        MIRACL~\cite{zhang2022making, multim3} & Retrieval & 11.0 & search\_query/search\_document \\
        Mr. Tydi~\cite{mrtydi, multim3} & Retrieval & 9.0 & search\_query/search\_document \\
        \bottomrule
    \end{tabular}
    \caption{The full training corpus with corresponding prefixes. We report the number of pairs in thousands. For the tasks with two different prompts, query and document, they are written with a slash.}
    \label{tab:dataset_info}
    \end{minipage}
\end{table*}

\subsubsection{Data Filtration Details}
\label{app:data_filtration}

We apply the following steps to the basic Russian datasets. First, texts longer than 500 tokens (ruRoBERTa-large\footnote{\url{https://huggingface.co/ai-forever/ruRoberta-large}} tokenizer is used) are filtered out. A small number of tokens is reserved for instructions or prefixes. Pairs from YandexQ\footnote{\url{https://huggingface.co/datasets/IlyaGusev/yandex_q_full}}, Pikabu\footnote{\url{https://huggingface.co/datasets/IlyaGusev/pikabu}}, StackOverflow\footnote{\url{https://huggingface.co/datasets/IlyaGusev/ru_stackoverflow}}, Habr\footnote{\url{https://huggingface.co/datasets/IlyaGusev/habr}} and Habr QnA\footnote{\url{https://huggingface.co/datasets/its5Q/habr_qna}} are filtered by content popularity (e.g. views, ratings, votes). Cosine similarity obtained from LaBSE~\cite{feng2022language} is applied to filter NewsCommentary and MultiParaCrawl. We filter pairs from paraphrase-NMT-Leipzig\footnote{\url{https://huggingface.co/datasets/cointegrated/ru-paraphrase-NMT-Leipzig}} by \texttt{p\_good} score (equivalent meaning). The XNLI is formed from entailment (relevant document) and contradiction (irrelevant negative) examples. For MIRACL, we use the title as the query and the first paragraphs (until we reach the token limit) as the document. We form pairs for Paraphrases~\footnote{\url{https://huggingface.co/datasets/inkoziev/paraphrases}} from \texttt{paraphrases} field, taking one as a query and the others as positive documents. The content of RuNews\footnote{\url{https://huggingface.co/datasets/IlyaGusev/ru\_news}} is not changed. After exact match deduplication, the final training pairs for all datasets are randomly sampled from the remaining pairs.

\subsection{Model Training Details}
\label{app:training_details}

\subsubsection{Default training details}
We fine-tune the model in \texttt{bf16 dtype} with gradient checkpointing and use AdamW~\cite{loshchilov2017decoupled} with a learning rate of 1e-5 and weight decay of 0.01 for exactly one epoch, which is approximately 3700 steps, of which linear warmup is 200 steps. After fine-tuning, the SLERP merging is applied to the base model with a factor of 0.1.

We apply stratified sampling per device batch (mini-batch). Therefore, the global batch includes mini-batches consisting of pairs of different datasets. On the one hand, it becomes impossible to exchange negatives between devices and thus scale the number of in-batch negatives. On the other hand, this increases the diversity of sources in the global batch. Therefore, we do not apply the DisCo~\cite{chen2023discoclip} trick to exchange negative examples across devices. The batch size is 128 per device, giving 1024 documents per query. The context length is set to 512 for queries and documents~\cite{merrick2024arctic}.

Training is conducted on a single H100 node. We utilize the BGE\footnote{\url{https://github.com/FlagOpen/FlagEmbedding}} codebase and adapt it to our experiments. PyTorch's \verb|expandable_segments| helps us to mitigate fragmentation issues due to variable sequence length.

\subsubsection{Ablation training details}
\label{app:ablation_training}

\vspace{0.5em} \noindent\textbf{Remove prefixes}. We omit prefixes and keep the training process unchanged, preventing the model from identifying task types during training and inference.

\vspace{0.5em} \noindent\textbf{Disable stratified sampling}. Batch examples are randomly selected from all datasets instead of the single source. The prefixes are used only on the query side; otherwise the objective becomes easy to solve since different datasets have their own prefixes. DisCo is enabled to increase the number of negatives per query, unlike in stratified sampling.

\vspace{0.5em} \noindent\textbf{Hard negatives}. The number of hard negatives per query is increased from 7 to 15, while the batch size is reduced from 128 to 64 to maintain the same total number of negatives, and the gradient step accumulation (2 steps) is applied to keep training steps consistent.

\subsection{ruMTEB Dataset Description}\label{app:data_prep}

This section describes new tasks we present with the research and data preparation details.

\subsubsection{Classification}

\vspace{0.5em} \noindent\textbf{KinopoiskSentimentClassification}. In a sentiment classification dataset given a film review, one has to predict whether it is Positive, Neutral, or Negative (3 classes in total). The data was taken from the original dataset ~\cite{blinov2013research}\footnote{\url{https://huggingface.co/datasets/blinoff/kinopoisk}}, which contains reviews from July 2004 to November 2012. In the preprocessing phase, we removed all mentions of the final rating from the review texts and balanced the set, leaving only 4,500 samples of each class. The resulting dataset was split into three parts (train, validation, and test), with the class balance preserved.

\vspace{0.5em} \noindent\textbf{GeoReviewClassification}.
A classification dataset, where given a review text one has to predict its rating ranging from 1 to 5 (five classes in total). The set is based on the Yandex Maps\footnote{\url{https://yandex.ru/maps}} reviews\footnote{\url{https://github.com/yandex/geo-reviews-dataset-2023}}. The original dataset was balanced and split into three parts (train, validation, and test).

\vspace{0.5em} \noindent\textbf{HeadlineClassification}. In this dataset, the model needs to determine which news category the article title belongs to. The dataset was built based on ParaPhraserPlus~\cite{gudkov2020automatically} and contained 10,000 examples for each category, divided into train/validation/test splits of 6000, 2000, and 2000, respectively. A total of 6 classes are used: sports, incidents, politics, science, culture and economics. First, categories that contained at least 10,000 examples were selected. Other categories were discarded due to overlap between categories. For this purpose, we trained a classifier over SBERT$_{\textnormal{\scriptsize large-nlu-ru}}$ embeddings.

\vspace{0.5em} \noindent\textbf{RuReviewsClassification}.
A sentiment classification dataset where top-ranked goods from a major e-commerce site were provided, and user-ranked scores were used as class labels on a 5-point scale. The data was sourced from the original dataset RuReviews\footnote{\url{https://github.com/sismetanin/rureviews?tab=readme-ov-file}}, which contains reviews in the ``Women’s Clothes and Accessories'' category. During the preprocessing stage, duplicates were removed, and the dataset was balanced, resulting in only 25,000 samples for each class. The resulting dataset was divided into three parts (train, validation, and test) while maintaining class balance.

\vspace{0.5em} \noindent\textbf{RuSciBenchGRTNI/OECDClassification}.
This is a dataset for the classification of scientific text headings. Each article has its OECD and GRNTI headings, with 29 OECD headings and 28 GRNTI headings in the dataset (e.g., Mathematics, Biological Sciences, Economics and Business, etc.). The data was sourced from the original dataset RuSciBench\footnote{\url{https://huggingface.co/datasets/mlsa-iai-msu-lab/ru_sci_bench}}. During preprocessing, duplicates were removed, the title and abstract were combined, and the set was balanced, leaving only the same number of samples for each class. The resulting dataset was then divided into test and training parts.

\vspace{0.5em} \noindent\textbf{InappropriatnessClassification}. The dataset aims to predict whether the message is inappropriate or not in the form of binary classification. The data is based on the Inappropriate Messages dataset (version 3)\footnote{\url{https://github.com/s-nlp/inappropriate-sensitive-topics/blob/main/Version3/Inappapropriate_messages.csv}}~\cite{babakov-etal-2021-detecting}. We binarized the inappropriateness scores using the 0.5 threshold. The resulting dataset was balanced and split into three parts (train, validation, and test), with the class balance preserved.

\subsubsection{Pair Classification}

\vspace{0.5em} \noindent\textbf{TERRa}. The dataset was presented as one of the Russian SuperGlue tasks~\cite{shavrina2020russiansuperglue} and related to the Textual Entailment Recognition task. Given two texts, the task is to determine whether the meaning of one text entailed from the another text. Since the test split is hidden, we took the dev split without changes. A total of 307 examples are available.

\subsubsection{Multi-Label Classification}

\vspace{0.5em} \noindent\textbf{CEDRClassification}. The dataset is a task of classifying comments into five emotions (joy, sadness, surprise, fear, and anger). A total of 9,410 comments were presented from the following sources: social networks, news, and blogs. The dataset was used as is, without any modifications~\cite{sboev2021data}. We took the original test split, which includes 1882 examples.

\vspace{0.5em} \noindent\textbf{SensitiveTopicsClassification}. The dataset contains sentences that can be classified into one or more sensitive topics\footnote{\url{https://github.com/s-nlp/inappropriate-sensitive-topics/blob/main/Version3/sensitive_topics.csv}}~\cite{babakov-etal-2021-detecting}. The original dataset includes 18 classes, all classes are used. Since part of the dataset is not only manually labeled, we first formed a test split from manually labeled examples, and the remaining examples were combined with semi-automatically labeled examples. We have selected the most reliable examples based on the confidence scores indicated in the examples. The final test split consists of 2048 examples and preserves the original class distribution.

\subsubsection{Clustering}

\vspace{0.5em} \noindent\textbf{GeoReviewClustering}.
A clustering dataset based on the Yandex Maps\footnote{\url{https://yandex.ru/maps}} reviews\footnote{\url{https://github.com/yandex/geo-reviews-dataset-2023}}, where given a review text one has to cluster the samples according to their rubrics or review categories (e.g., Bank, Supermarket, Pharmacy, etc.). The original dataset was balanced and split into three parts (train, validation, and test). For each review, we took its first rubric as the main label, leaving only samples corresponding to the top 100 most popular labels. This threshold limited the categories exceeding 10,000 examples. The final dataset was converted into the MTEB format.

\vspace{0.5em} \noindent\textbf{RuSciBenchGRTNI/OECDClustering}.
This is a dataset for the clustering of scientific text headings. Each article has its OECD and GRNTI headings, and there are 29 OECD headings and 28 GRNTI headings in the dataset (e.g., Mathematics, Biological Sciences, Economics and Business, etc.). The data was sourced from the original dataset RuSciBench\footnote{\url{https://huggingface.co/datasets/mlsa-iai-msu-lab/ru\_sci\_bench}}. During preprocessing, duplicates were removed, the title and abstract were combined, and the set was balanced, leaving only the same number of samples for each class. The resulting dataset was then divided into test and training parts.

\subsubsection{Semantic Textual Similarity (STS)} 

\vspace{0.5em} \noindent\textbf{RuSTSBenchmarkSTS}. The dataset used for the STS task is derived from the original multilingual STS Benchmark~\footnote{\url{https://github.com/PhilipMay/stsb-multi-mt}}. This multilingual set comprises various translations of the original English version of the STSbenchmark dataset, with the translations completed using deepl.com~\footnote{\url{https://www.deepl.com/ru/translator}}. The Russian segment of the dataset was extracted and refined using the RuCoLa~\cite{Mikhailov_2022} classifier~\footnote{\url{https://huggingface.co/RussianNLP/ruRoBERTa-large-rucola}}. In all parts of the sets (train/dev/test), instances categorized as not linguistically acceptable were excluded. Additionally, any duplicate entries were eliminated.

\subsubsection{Reranking}

\vspace{0.5em} \noindent\textbf{RuBQReranking}. The dataset is based on RuBQ version 2.0~\cite{rybin2021rubq}. The dataset contains examples of questions and paragraphs from Wikipedia. Paragraphs that answer the question are considered relevant. Paragraphs that contain the answer are used as positive documents. Negative documents are paragraphs relevant to the question's topic but not the answer. We only used questions from the test split with at least nine negative documents. The final test split contains 1551 examples.

\subsubsection{Retrieval}

\vspace{0.5em} \noindent\textbf{RuBQRetrieval}. Unique paragraphs from the dataset are used for the document bank, resulting in 56,826 documents. Documents were deduplicated while links to relevant documents were maintained. The original test split was taken without changes and has 2845 examples.

\vspace{0.5em} \noindent\textbf{RiaNewsRetrieval}. The original dataset RussiaSegodnya\footnote{\url{https://github.com/RossiyaSegodnya/ria_news_dataset}} (also known as RiaNews) consists of news articles and their headlines~\cite{gavrilov2018self}. Texts are presented in lowercase format, and the capitalization of individual characters has not been changed. Since the article texts are available in HTML, we used the BeautifulSoup\footnote{\url{https://www.crummy.com/software/BeautifulSoup}} library to clean them of markup. Additionally, texts were normalized, and extra spaces were removed. We also removed, if possible, the first sentence in each article text since it does not relate to the article's content and is a kind of meta information (``Moscow, 1 Dec — ria news.''). We filtered out the texts of articles with more than 2000 characters so that models limited to a context of 512 tokens could handle the entire text. All examples were deduplicated based on the headline and text of the article. Our final dataset consists of 10,000 randomly sampled headlines as queries, and article texts (724344) are used as documents.

\subsection{Experimental Setup Details}
\label{app:prompts}
This section describes the prompt and embedding configuration we used in our experiments. Namely, we use normalized embeddings for evaluation on all ruMTEB tasks. We use pooling and instruction strategies required by the corresponding model we evaluate. Table~\ref{tab:prompts_rumteb} presents all the prefixes and instructions used. Specifically:
\begin{itemize}
    \item we do not utilize any special prompts for rubert-tiny2, BGE-M3, SBERT$_{\textnormal{\scriptsize large-nlu-ru}}$, and SBERT$_{\textnormal{\scriptsize large-mt-nlu-ru}}$;
    \item we use special prefixes for ru-en-RoSBERTa;
    \item mE5$_{\textnormal{\scriptsize small/medium/large}}$ models share the same set of prefixes;
    \item mE5$_{\textnormal{\scriptsize large-instruct}}$ and E5$_{\textnormal{\scriptsize mistral-7b-instruct}}$ models share the same set of instructions.
\end{itemize}

\begin{table*}[h]
    \centering
    \tiny
    \begin{tabular}{*{4}{l}}
        \toprule
        \textbf{Task name}   & \textbf{ru-en-RoSBERTa} & \textbf{E5 prefix} & \textbf{E5 instruction} \\
        \midrule
        \textit{Classification} \\
        \midrule
        GeoreviewClassification & classification & query & Classify the organization rating based on the reviews \\
        HeadlineClassification & clustering & query & Classify the topic or theme of the given news headline \\
        InappropriatenessClassification & clustering & query & Classify the given message as either sensitive topic or not \\
        KinopoiskClassification & classification & query & Classify the sentiment expressed in the given movie review text \\
        MassiveIntentClassification & classification & query & Given a user utterance as query, find the user intents \\
        MassiveScenarioClassification & clustering & query & Given a user utterance as query, find the user scenarios \\
        RuReviewsClassification & classification & query & Classify product reviews into positive, negative or neutral sentiment \\
        RuSciBenchGRNTIClassification & clustering & query & Classify the category of scientific papers based on the titles and abstracts \\
        RuSciBenchOECDClassification & clustering & query & Classify the category of scientific papers based on the titles and abstracts \\
        \midrule
        \textit{Clustering} \\
        \midrule
        GeoreviewClusteringP2P & clustering & query & Identify the organization category based on the reviews \\
        RuSciBenchGRNTIClusteringP2P & clustering & query & Identify the category of scientific papers based on the titles and abstracts \\
        RuSciBenchOECDClusteringP2P & clustering & query & Identify the category of scientific papers based on the titles and abstracts \\
        \midrule
        \textit{MultiLabelClassification} \\
        \midrule
        CEDRClassification & classification & query & Given a comment as query, find expressed emotions (joy, sadness, surprise, fear, and anger) \\
        SensitiveTopicsClassification & clustering & query & Given a sentence as query, find sensitive topics \\
        \midrule
        \textit{PairClassification} \\
        \midrule
        TERRa & classification & query & Given a premise, retrieve a hypothesis that is entailed by the premise \\
        \midrule
        \textit{Reranking} \\
        \midrule
        MIRACLReranking & search\_query/search\_document & query/passage & Given a question, retrieve Wikipedia passages that answer the question \\
        RuBQReranking & search\_query/search\_document & query/passage & Given a question, retrieve Wikipedia passages that answer the question \\
        \midrule
        \textit{Retrieval} \\
        \midrule
        MIRACLRetrieval & search\_query/search\_document & query/passage & Given a question, retrieve Wikipedia passages that answer the question \\
        RiaNewsRetrieval & search\_query/search\_document & query/passage & Given a news title, retrieve relevant news article \\
        RuBQRetrieval & search\_query/search\_document & query/passage & Given a question, retrieve Wikipedia passages that answer the question \\
        \midrule
        \textit{STS} \\
        \midrule
        RUParaPhraserSTS & classification & query & Retrieve semantically similar text \\
        RuSTSBenchmarkSTS & classification & query & Retrieve semantically similar text \\
        STS22 & clustering & query & Retrieve semantically similar text \\
        \bottomrule
    \end{tabular}
    \caption{Prompts used for ruMTEB evaluation. For the tasks with two different prompts, query and document, they are written with a slash. \textit{E5 prefix} shows prefixes used for mE5$_{\textnormal{\footnotesize small/medium/large}}$ models. \textit{E5 instruction} shows instructions used for E5$_{\textnormal{\footnotesize mistral-7b-instruct}}$ and mE5$_{\textnormal{\footnotesize large-instruct}}$ models.}
    \label{tab:prompts_rumteb}
\end{table*}

\subsection{Detailed Results}
\label{app:all_results}
Table~\ref{tab:rumteb_task_results} shows results on individual ruMTEB datasets. We run evaluation on NVIDIA A100 80GB with torch 2.2.1+cu118 and transformers 4.40.2. Please refer to PR\footnote{\url{https://github.com/embeddings-benchmark/results/pull/19}} to access the results.

\begin{table*}[h]
    \tiny
    \centering
    \begin{tabular}{l*{10}{c}}
\toprule
\multirow{2}{*}{ } & rubert & SBERT & SBERT & mE5 & mE5 & mE5 & BGE-M3 & ru-en-RoSBERTa & mE5 & E5 \\
             & tiny2 & large-nlu-ru & large-mt-nlu-ru & small & base & large & & & large-instruct & mistral-7b-instruct \\
\midrule
\textit{Classification} \\
\midrule
GeoreviewClassification         & 39.64             & 39.97             & 39.67             & 44.66             & 46.05             & 49.69             & 48.27             & 49.70             & \underline{55.90}             & \textbf{56.72}             \\
HeadlineClassification          & 74.19             & 79.26             & 77.19             & 73.94             & 75.64             & 77.19             & 70.32             & 78.00             & \underline{86.18}             & \textbf{87.02}             \\
InappropriatenessClassification & 58.57             & 62.52             & 64.64             & 59.16             & 58.78             & 61.59             & 59.87             & 61.32             & \underline{65.53}             & \textbf{70.36}             \\
KinopoiskClassification         & 49.06             & 49.51             & 50.33             & 49.96             & 50.89             & 56.59             & 58.23             & 63.27             & \underline{66.12}             & \textbf{68.35}             \\
MassiveIntentClassification     & 50.83             & 61.09             & 61.42             & 58.43             & 62.78             & 65.76             & \underline{68.76}             & 66.97             & 67.60             & \textbf{73.74}             \\
MassiveScenarioClassification   & 59.15             & 67.60             & 68.13             & 63.89             & 68.21             & 70.85             & \underline{73.42}             & 71.80             & 71.59             & \textbf{77.10}             \\
RuReviewsClassification         & 56.99             & 58.27             & 58.29             & 61.18             & 62.99             & 65.28             & 66.91             & 67.96             & \underline{68.56}             & \textbf{70.57}             \\
RuSciBenchGRNTIClassification   & 45.63             & 53.90             & 54.19             & 54.99             & 56.28             & 58.20             & 55.81             & 59.33             & \underline{65.07}             & \textbf{66.05}             \\
RuSciBenchOECDClassification    & 35.48             & 43.04             & 43.80             & 41.72             & 42.69             & 43.91             & 42.57             & 46.33             & \underline{50.21}             & \textbf{52.11}             \\
\midrule
\textit{Clustering} \\
\midrule
 GeoreviewClusteringP2P          & 41.58             & 57.12             & 57.07             & 58.57             & 54.46             & 59.59             & 63.09             & 65.42             & \underline{74.34}             & \textbf{76.32}             \\
 RuSciBenchGRNTIClusteringP2P    & 39.78             & 49.70             & 51.44             & 51.14             & 51.56             & 51.98             & 50.83             & 55.47             & \underline{62.21}             & \textbf{62.27}             \\
 RuSciBenchOECDClusteringP2P     & 35.98             & 44.48             & 45.36             & 44.33             & 44.79             & 45.12             & 43.21             & 47.29             & \underline{53.09}             & \textbf{54.13}             \\
\midrule
\textit{MultiLabelClassification} \\
\midrule
CEDRClassification              & 36.87             & 35.84             & 36.81             & 40.07             & 42.32             & 44.84             & 43.47             & 44.69             & \underline{50.01}             & \textbf{51.94}             \\
SensitiveTopicsClassification   & 22.03             & 27.90             & 28.54             & 23.91             & 24.98             & 27.17             & 26.25             & \underline{33.07}             & 32.29             & \textbf{33.92}             \\
\midrule
\textit{PairClassification} \\
\midrule
TERRa                           & 51.87             & 50.17             & 51.97             & 55.14             & 54.98             & 58.42             & 60.60             & 60.79             & \textbf{63.89}             & \underline{60.81}             \\
\midrule
\textit{Reranking} \\
\midrule
MIRACLReranking                 & 15.81             & 18.80             & 24.99             & 59.11             & 60.47             & \underline{63.71}            & \textbf{65.38}             & 56.91             & 62.49             & 63.61             \\
RuBQReranking                   & 46.09             & 46.81             & 56.14             & 71.45             & 72.01             & 75.60             & 74.03             & 70.87             & \underline{75.84}             & \textbf{76.32}             \\
\midrule
\textit{Retrieval} \\
\midrule
MIRACLRetrieval                 & 1.89              & 1.98              & 6.20              & 59.01             & 61.60             & 67.33             & \textbf{70.16}             & 53.91             & 66.08             & \underline{67.66}             \\
RiaNewsRetrieval                & 13.92             & 11.11             & 21.40             & 70.00             & 70.24             & 80.67             & \underline{82.99}             & 78.86             & \textbf{83.26}             & 78.94             \\
RuBQRetrieval                   & 10.87             & 12.45             & 29.80             & 68.53             & 69.58             & \underline{74.13}             & 71.22             & 66.77             & 73.90             & \textbf{75.98}             \\
\midrule
\textit{STS} \\
\midrule
RUParaPhraserSTS                & 65.14             & 62.06             & 65.17             & 70.46             & 70.17             & 71.82             & 74.90             & \underline{76.16}             & 75.40             & \textbf{76.17}             \\
RuSTSBenchmarkSTS               & 69.43             & 58.82             & 71.22             & 78.08             & 79.64             & 83.15             & 79.87             & 78.69             & \underline{83.97}             & \textbf{84.13}             \\
STS22                           & 50.23             & 50.75             & 56.82             & 59.90             & 60.67             & 59.89             & \underline{66.26}             & \textbf{67.06}             & 65.17             & 60.83             \\
\midrule
Average                         & 42.22             & 45.35             & 48.72             & 57.29             & 58.34             & 61.41             & 61.58             & 61.77             & \underline{66.03}             & \textbf{67.18}             \\
\bottomrule
\end{tabular}
\caption{The full results of model evaluation on the ruMTEB benchmark. The aggregated score for each task category is reported in Section~\ref{sec:results}. The best
score is put in bold, the second best is underlined.}\label{tab:rumteb_task_results}
\end{table*}

\subsection{Additional Experimental Findings}
\label{app:additional_notes}

In this part, we describe early-stage experiments that were conducted on different data subsets and different base models.

\vspace{0.5em} \noindent\textbf{Prefixes}. We found that the E5 prefixes~\cite{wang2022text} performed slightly worse and assume that the variant we use helps to better separate tasks during training. The \verb|clustering| prefix is more suitable for tasks where thematic identification is required, so in many classification problems, we use it instead of \verb|classification|, despite the name. We tried adding prefixes with some probability; this improved the results without using prefixes and also worsened the results with them. In addition to stratified sampling, we implemented a sampling strategy that takes pairs with the same prefix but saw no improvement.

\vspace{0.5em} \noindent\textbf{Losses}. It was shown that Sigmoid Loss~\cite{zhai2023sigmoid} performed better at smaller batch sizes. We found that SigLIP is more sensitive to selecting the initial values of the bias and temperature parameters to achieve convergence. CoSENT loss~\cite{li2023angle} shows better results for STS-like tasks; we adapted the loss for the case with many negatives. In both cases, we were unable to achieve comparable results and left this for further work.

\vspace{0.5em} \noindent\textbf{Augmentations}. Although the model trained on 1500 steps shows comparable results to full training, we tried to apply text level and embedding level augmentations but found no meaningful performance improvement. For the text level, we used character-level augmentation from the Augmentex\footnote{\url{https://github.com/ai-forever/augmentex}} library~\cite{martynov2024methodology} for both languages. In another experiment, we applied the NEFTune~\cite{jain2023neftune} with 3, 5, and 10 alpha parameters.

\end{document}